\documentclass[10pt,twocolumn,letterpaper]{article}

\usepackage[pagenumbers]{wacv}      

\usepackage[T1]{fontenc}
\usepackage{subcaption}
\usepackage{graphicx}

\usepackage{tabularx}
\usepackage{xcolor}
\usepackage{listings}
\usepackage{tcolorbox}
\usepackage{standalone}

\definecolor{eclipseStrings}{RGB}{42,0.0,255}
\definecolor{eclipseKeywords}{RGB}{127,0,85}
\colorlet{numb}{magenta!60!black}

\lstdefinelanguage{json}{
    basicstyle=\normalfont\ttfamily\footnotesize,
    numbers=left,
    numberstyle=\scriptsize,
    stepnumber=1,
    numbersep=8pt,
    showstringspaces=false,
    breaklines=true,
    frame=lines,
    backgroundcolor=\color{gray!5},
    stringstyle=\color{eclipseStrings},
    literate=
     *{0}{{{\color{numb}0}}}{1}
      {1}{{{\color{numb}1}}}{1}
      {2}{{{\color{numb}2}}}{1}
      {3}{{{\color{numb}3}}}{1}
      {4}{{{\color{numb}4}}}{1}
      {5}{{{\color{numb}5}}}{1}
      {6}{{{\color{numb}6}}}{1}
      {7}{{{\color{numb}7}}}{1}
      {8}{{{\color{numb}8}}}{1}
      {9}{{{\color{numb}9}}}{1}
      {:}{{{\color{black}:}}}{1}
      {,}{{{\color{black},}}}{1}
      {\{}{{{\color{black}\{}}}{1}
      {\}}{{{\color{black}\}}}}{1}
      {[}{{{\color{black}[}}}{1}
      {]}{{{\color{black}]}}}{1},
}

\usepackage{float}
\usepackage{tcolorbox}
\usepackage{caption}

\floatstyle{plain} 
\newfloat{promptfloat}{H}{lop}
\floatname{promptfloat}{Prompt}

\newtcolorbox{promptbox}{
  colback=gray!5,
  colframe=gray!50,
  boxrule=0.5mm,
  arc=2mm,
  left=5pt, right=5pt, top=5pt, bottom=5pt,
  fontupper=\small\ttfamily
}

\definecolor{wacvblue}{rgb}{0.21,0.49,0.74}
\usepackage[pagebackref,breaklinks,colorlinks,allcolors=wacvblue]{hyperref}

\def\wacvPaperID{589} 
\def\confName{WACV}
\def\confYear{2027}

\title{VIGIL: Tackling Hallucination Detection in Image Recontextualization}

\author{Joanna Wojciechowicz$^{*1}$ \quad 
Maria Łubniewska$^{*1}$ \quad 
Wojciech Gromski$^{*1}$ \quad 
Jakub Antczak$^{*1}$ \\
Justyna Baczyńska$^{*1}$ \quad 
Wojciech Maciej Kozłowski$^{1}$ \quad 
Maciej Zięba$^{1,2}$ \\
\\
$^{1}$Wroclaw University of Science and Technology \quad $^{2}$Tooploox \\
{\tt\small \{255747, 268845, 268725, 268745, 268758\}@student.pwr.edu.pl} \\
{\tt\small \{wojciech.kozlowski, maciej.zieba\}@pwr.edu.pl}
}

\begin{document}
\makeatletter
\twocolumn[{%
  \@maketitle
  \vspace{-18pt}
  \centering
  \includegraphics[width=\textwidth]{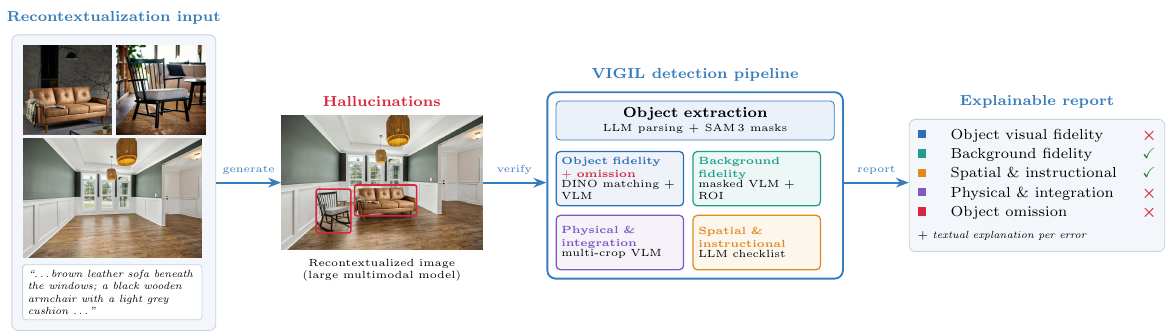}%
  \captionof{figure}{\textbf{Overview of VIGIL.} Given reference object images, a target background, and a textual prompt, a large multimodal model recontextualizes the objects into the scene but may introduce hallucinations. VIGIL first extracts the prompted objects (LLM parsing and SAM~3) and then verifies the result with open-source models across all five hallucination categories of our taxonomy---object visual fidelity, object omission, background fidelity, physical and integration fidelity, and spatial and instructional fidelity. Unlike prior detectors that emit a single opaque score, it returns a per-category verdict together with a textual explanation of each detected error.}%
  \label{fig:teaser}
  \vspace{1.2em}%
}]
\makeatother
\thispagestyle{empty}

\let\thefootnote\relax\footnotetext{$^*$Equal contribution}

\begin{abstract}
We introduce VIGIL (\textbf{V}isual \textbf{I}nconsistency \& \textbf{G}enerative \textbf{I}n-context \textbf{L}ucidity), a benchmark dataset and framework that provides a fine-grained categorization of hallucinations in the multimodal image recontextualization task for large multimodal models. Most existing methods treat hallucinations as a single undifferentiated error. We instead decompose them into five categories, namely Object Visual Fidelity, Background Fidelity, Spatial and Instructional Fidelity, Physical and Integration Fidelity, and Object Omission. We propose a multi-stage detection pipeline that processes recontextualized images through specialized steps targeting all five categories with a coordinated set of open-source models. We evaluate the pipeline on the VIGIL dataset of 1,269 manually annotated samples across five product domains and report macro-F1 per category. The decomposed pipeline reaches the best macro-F1 among open-source detectors. It returns a textual explanation for each detected error, which prior methods for this task do not provide. The code and dataset will be made publicly available.

\end{abstract}
    
\section{Introduction}
\label{sec:intro}

Conditional image generation has advanced rapidly through two paradigms. Text-to-Image (T2I) synthesis generates imagery from textual descriptions (e.g., Flux.1 \cite{labs2025flux1kontextflowmatching}, DALL-E 3 \cite{ramesh2021zeroshottexttoimagegeneration}). Image recontextualization synthesizes new scenes by combining text with specific visual subjects from reference images. Multimodal systems such as Gemini 3 Pro \cite{geminiteam2025geminifamilyhighlycapable}, 
GPT-5.2, 
and the Qwen3.5 series \cite{qwen3.5technicalreport} produce subject-driven generations with high realism. Even these models frequently produce hallucinations, that is visual artifacts, identity shifts, and semantic inconsistencies that have no support in the input prompts or reference images. Such hallucinations reduce the reliability of generation systems in applications like automated advertising and digital content creation, where the attributes of a reference product must be preserved. Hallucination detection has been studied mainly within the T2I domain, where it focuses on the alignment between text prompts and generated pixels. Some recent methods extend this evaluation to multimodal inputs (\texttt{text + images}), but they report coarse numerical scores with limited interpretability. These scores rarely give descriptive feedback or distinguish error types such as identity distortion from environmental mismatch. The absence of a standardized, fine-grained evaluation framework limits the diagnostic analysis needed to move from research to production use.

We address this gap with an evaluation framework and benchmark dataset for hallucination assessment in multimodal recontextualization. Our pipeline uses SAM 3 \cite{carion2025sam3segmentconcepts}, DINO v3 \cite{simeoni2025dinov3}, and Qwen3.5 \cite{qwen3.5technicalreport} to produce human-interpretable, descriptive feedback rather than a single score. We evaluate the approach on a dataset generated with Gemini 2.5 Flash Image, with manual annotations that capture the ways current generative models fail.

Our primary contributions are the following.
\begin{itemize}
    \item We present a dataset of 1,269 samples generated with Gemini 2.5 Flash Image. The benchmark includes manual annotations across five categories of hallucinations and provides ground truth for multimodal research.
    \item We propose a method that automatically detects and describes five categories of hallucinations. The decomposed pipeline matches or exceeds open-source Vision-Language Model (VLM) detectors and approaches a much larger proprietary model, while it returns a textual explanation for each error rather than an isolated score.
    \item We define a taxonomy of hallucinations for the recontextualization task, which supports the diagnosis of model reliability for use cases such as advertising and digital content creation.
\end{itemize}


\section{Related Work}
Hallucination detection in generative models has been approached from several perspectives that range from model-agnostic methods to task-specific techniques. We review methods that rely only on generated outputs, methods that analyze model behavior during generation, approaches based on input-output text--image alignment, and methods that target hallucinations in image recontextualization.\\

\noindent
\textbf{Model agnostic hallucination detection}
Some approaches detect hallucinations using only the generated outputs. I-HallA \cite{ihalla} applies GPT-4o-based visual question answering to verify the factual consistency of generated outputs and releases a benchmark with textbook-derived prompts, hallucinated images, and human-validated QA pairs. WISE \cite{niu2025wise} provides a benchmark for real-world knowledge evaluation and a WiScore metric that measures semantic consistency and realism beyond text--image matching. LEGION \cite{legion} combines a synthetic dataset, SynthScars, with a framework that detects pixel-level artifacts, generates explanations, and guides image refinement. The authors in \cite{kea} use knowledge graphs to compare model outputs with ground-truth data, providing interpretable explanations of hallucinations in text generation. \\

\noindent
\textbf{Inference time hallucination detection}
Some methods detect hallucinations by analyzing model behavior during generation. For diffusion models \cite{ho2020denoising,sohl2015deep}, the authors in \cite{aithal2024understanding} identify out-of-distribution samples caused by smooth approximations over discontinuous data manifolds and propose a trajectory-variance metric to detect hallucinations. Building on this, \cite{dynamic_guidance} uses generation-time adaptive sharpening along artifact-prone directions to prevent hallucinations while preserving valid interpolations. ESIDE \cite{eside} analyzes the intermediate timesteps of DDIM \cite{song2020denoising} inversion to detect artifacts in T2I and recontextualization models, providing explanations and benchmarks for evaluation. For LLMs, LapEigvals \cite{binkowski2025hallucination} examines attention graphs using spectral features to predict hallucinations, while MB-ICL \cite{mbicl} selects in-context learning examples based on latent manifolds to improve the detection of factual inconsistencies across tasks. \\

\noindent
\textbf{Hallucination detection in text-image alignment}
For text-to-image and image captioning models, hallucination detection focuses on identifying inconsistencies between text and images. To measure text--image alignment, POPE \cite{pope} uses Yes/No questions derived from generated captions and images from the MSCOCO dataset \cite{lin2014microsoft}.
CHAIR \cite{chair} measures the proportion of hallucinated objects among all objects mentioned in the text description. AMBER \cite{amber} improves on CHAIR by evaluating not only the existence but also the attributes and relationships between the objects.
LURE \cite{lure} and Woodpecker \cite{woodpecker} detect and correct hallucinations post-hoc, with LURE using co-occurrence patterns and uncertainty, and Woodpecker validating visual claims in multiple stages by first extracting key objects from the image and then querying a VLM with increasingly specific questions about each object. PA-ICVL \cite{paicvl} addresses pose-related and structural hallucinations in non-photorealistic output. The authors in \cite{qa-agents} use knowledge-based question answering on scene graphs, while REVEALER \cite{revealer} applies reinforcement learning and structured reasoning for element-level evaluation.
The most prominent datasets for the validation of the above methods are GenAI-Bench \cite{genai-bench}, which provides large-scale human evaluation with VQA-based scoring, and AlignBench \cite{alignbench}, which offers fully annotated image--caption pairs for detailed alignment assessment. \\

\noindent
\textbf{Image recontextualization hallucination detection}
Recent work has extended hallucination detection to image recontextualization and other multi-domain conditional generation tasks. VIEScore \cite{viescore} introduces an instruction-guided, training-free metric that uses multimodal LLMs to provide a task-aware, explainable evaluation with rationales in natural language. CIGEval \cite{cigeval} proposes an agentic evaluation framework that uses reasoning-based multimodal models to construct a dynamic, task-specific pipeline without any human-designed alignment or pipeline structure.
ImagenHub \cite{imagenhub} provides a unified library, dataset, and evaluation framework across multiple generation tasks, standardizing inference and human evaluation to measure semantic consistency, perceptual quality, and hallucinations in various models.

\begin{figure*}[ht!]
  \centering
  \begin{subfigure}[c]{0.24\linewidth}
    \centering
    \includegraphics[width=\linewidth]{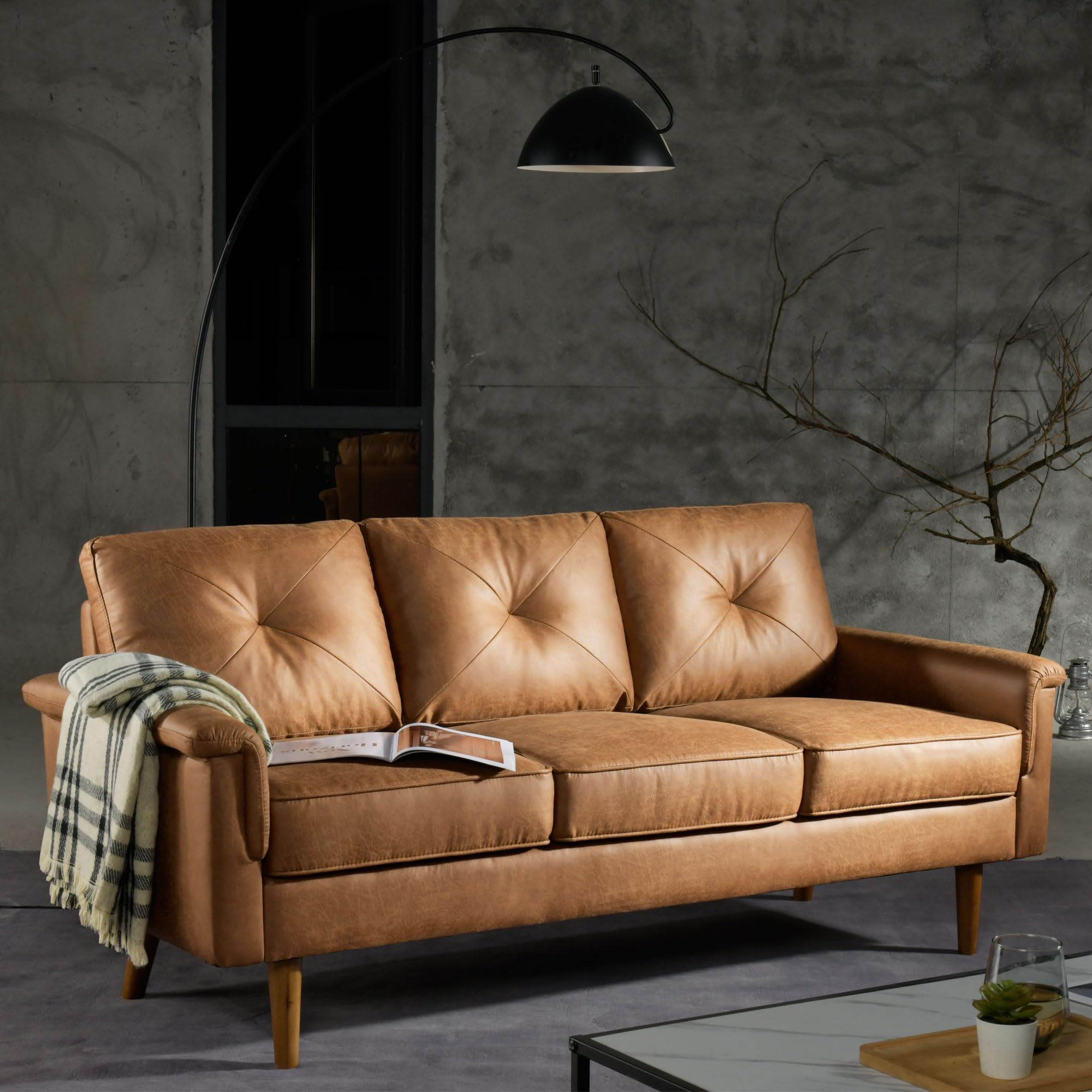} 
    \caption{Reference image of object 1.}
    \label{fig:object1_img}
  \end{subfigure}
  \hfill
  \begin{subfigure}[c]{0.24\linewidth}
    \centering
    \includegraphics[width=\linewidth]{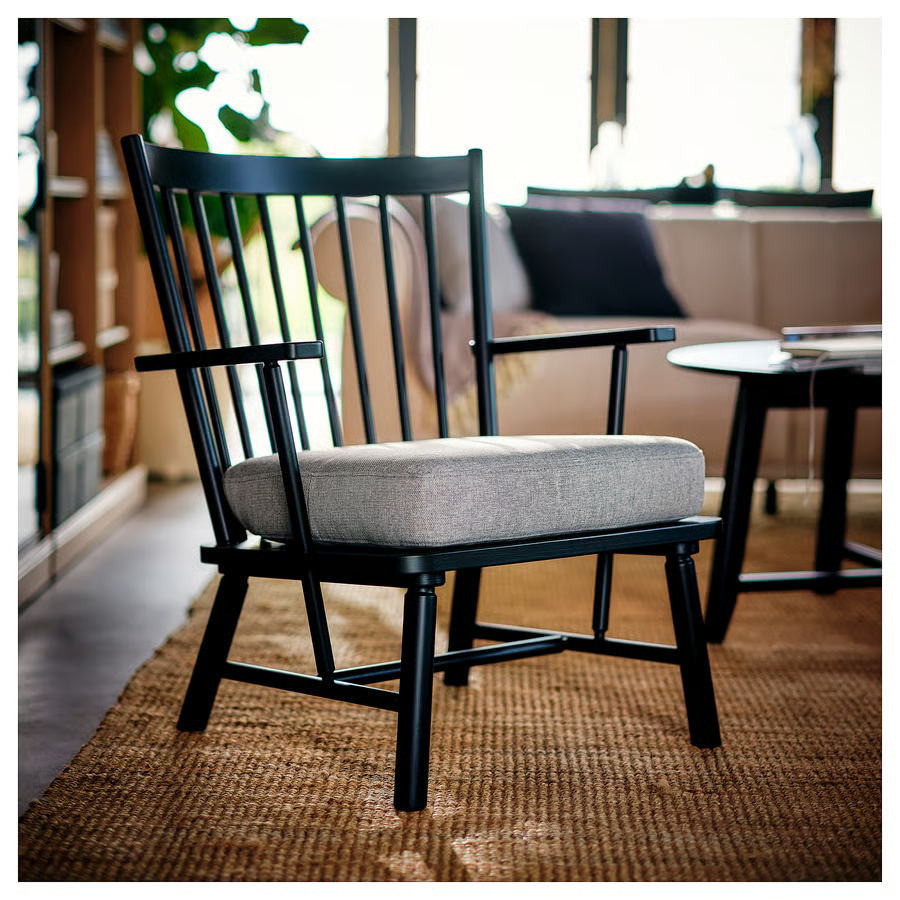}
    \caption{Reference image of object 2.}
    \label{fig:object2_img}
  \end{subfigure}
  \hfill
  \begin{subfigure}[c]{0.24\linewidth}
    \centering
    \includegraphics[width=\linewidth]{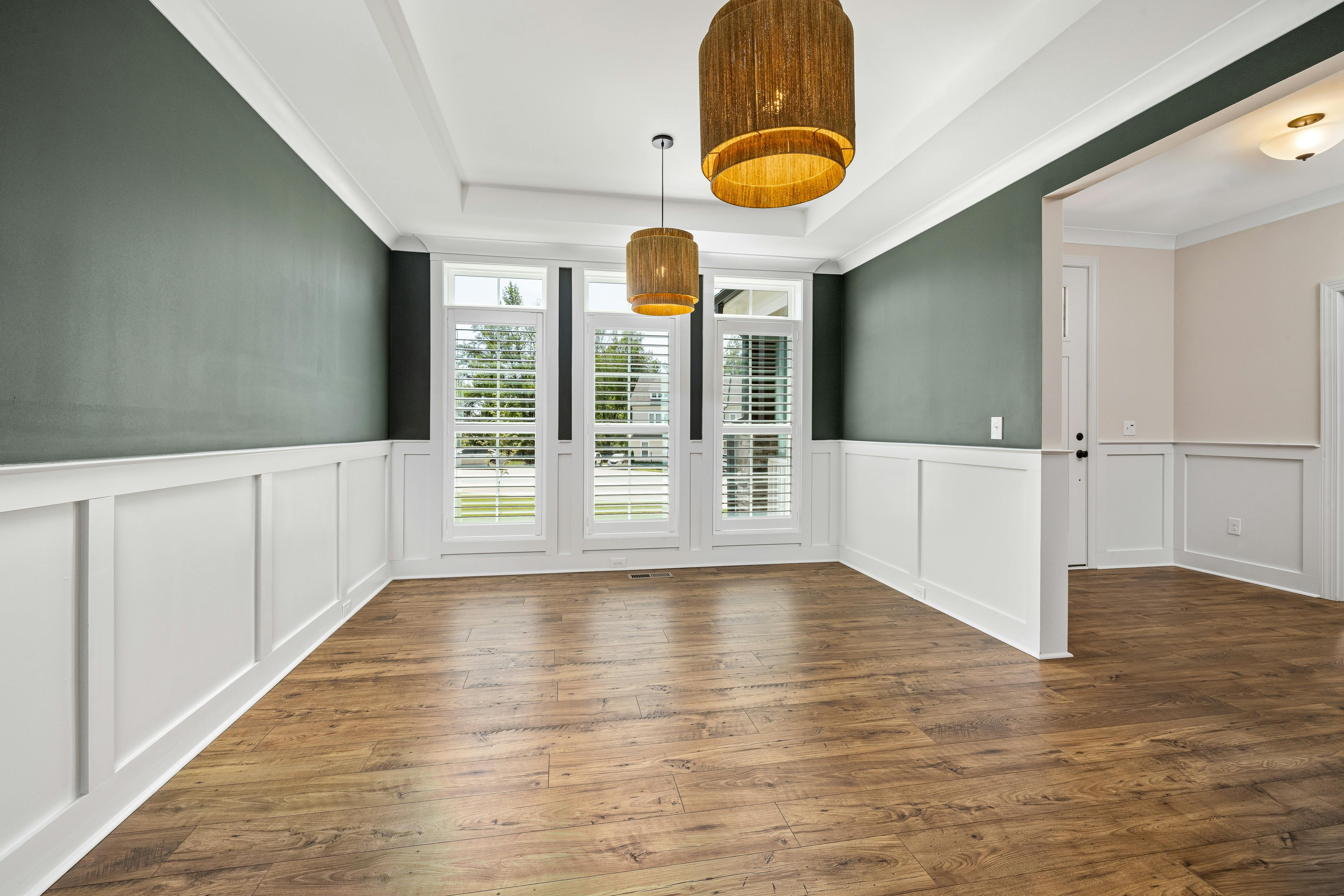}
    \caption{Background image.}
    \label{fig:background_img}
  \end{subfigure}
  \hfill
  \begin{subfigure}[c]{0.24\linewidth}
    \centering
    \includegraphics[width=\linewidth]{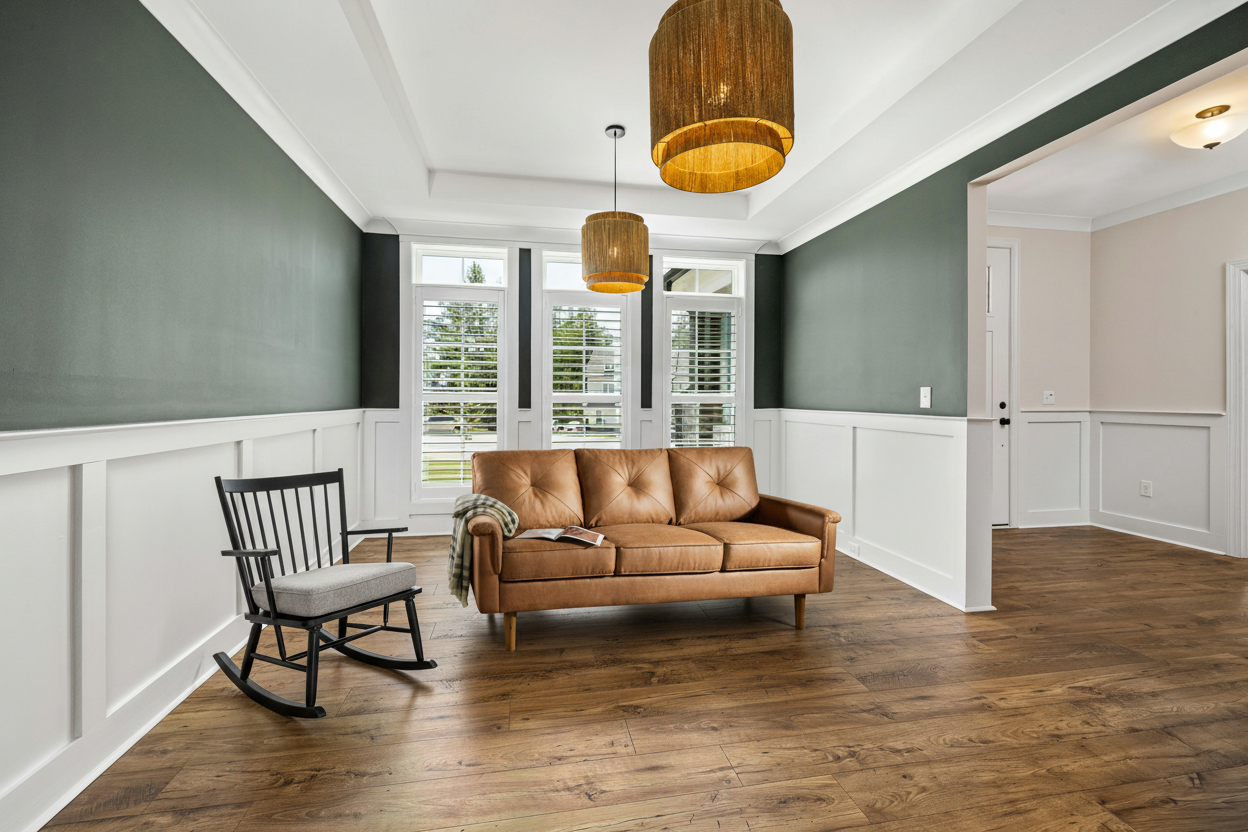}
    \caption{Generated image.}
    \label{fig:generated_img}
  \end{subfigure}
  \\
  \begin{promptbox}
    A photorealistic view of a spacious, modern dining room features dark green walls above white wainscoting, with three large windows centered on the far wall, letting in natural light; two woven rattan pendant lights hang from the ceiling. \textbf{A three-seater brown leather sofa, detailed with button tufting, an open book on its seat, and a plaid throw blanket draped over its left arm, is placed centrally against the far wall beneath the windows. In the midground, to the left of the sofa, a black wooden armchair with a light grey cushion is angled slightly towards the sofa.} Both pieces of furniture are realistically scaled and illuminated by the natural light from the windows, casting subtle shadows towards the viewer on the wooden floor, maintaining the original room's perspective and color harmony. No added text, no watermarks, no borders or frames, no unrelated objects.
  \end{promptbox}
  
  \caption{Exemplary sample from the VIGIL Benchmark Dataset. Each sample includes reference images for object 1 (\subref{fig:object1_img}) and object 2 (\subref{fig:object2_img}), the target background image (\subref{fig:background_img}), the final generated image (\subref{fig:generated_img}), and the corresponding textual recontextualization prompt used for generation.}
  \label{fig:full_sample_visualization}
\end{figure*}

\subsection{Gaps in Current Research}
Despite growing interest in hallucination detection, most methods focus on text--image alignment, and only a few recent approaches address multimodal recontextualization. However, these methods \cite{viescore,cigeval} typically quantify realism and alignment using simple scores, such as 0-10, and provide only unstructured textual descriptions of hallucinations. In contrast, our work introduces a detailed taxonomy of hallucinations and produces structured outputs that explicitly follow this taxonomy. Moreover, while there is a large existing benchmark \cite{imagenhub} for recontextualization, it is based on automatic annotations. Our dataset is smaller but manually annotated, ensuring higher-quality labels that follow the defined taxonomy and support more precise evaluation.
\section{Dataset}

\begin{figure}[b]
    \centering
    \includegraphics[width=0.4\textwidth]{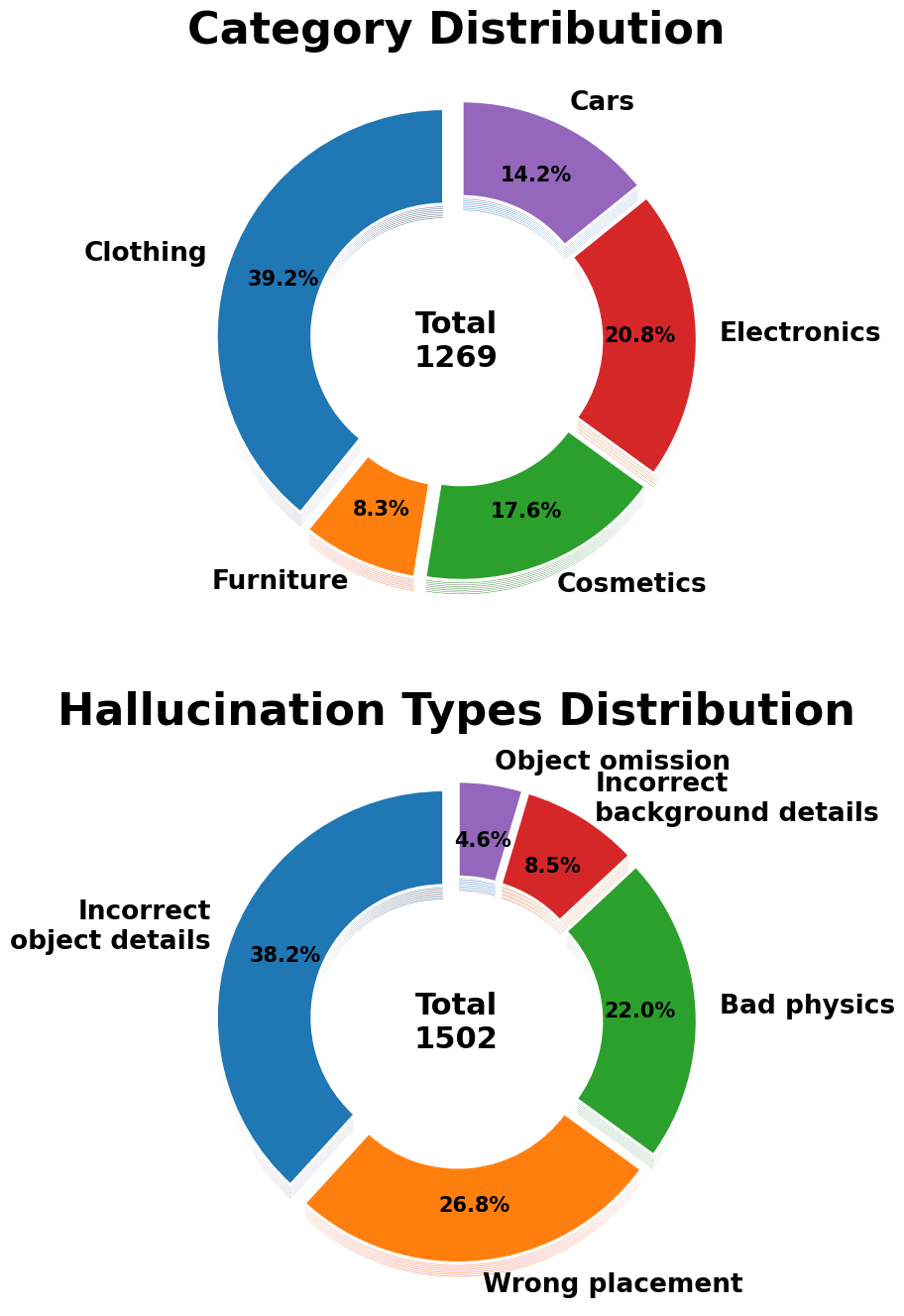}
    \caption{The first chart shows the distribution of the categories, while the second shows the proportion of different hallucination types considered in our framework.}
    \label{fig:piecharts}
\end{figure}

To address the lack of benchmarks for multimodal recontextualization, we constructed a dataset of 1,269 samples across five semantic categories, namely clothing, furniture, cosmetics, electronics, and cars. These domains reflect applications in marketing and visually guided product design. Figure~\ref{fig:piecharts} shows the composition. Each entry consists of a background image, reference images for one or two objects, the resulting recontextualized image, and the textual prompt. Each example also includes JSON-formatted metadata with the assigned hallucination classes and explanatory rationales (see Figures~\ref{fig:full_sample_visualization} and \ref{fig:annotation_json}). In the JSON \textit{objects} stands for  Object Visual Fidelity, \textit{background} for Background Fidelity, \textit{position\_logic} for  Spatial\& Instructional Fidelity, \textit{physical} for  Physical\& Integration Fidelity and \textit{object\_omission} for Object Omission category.

The dataset contains 1,024 images with hallucinations (80.7\%) and 245 clean reference samples (19.3\%). A single image may exhibit several hallucinations, so the annotation yields 574 Object Visual Fidelity errors, 402 Spatial \& Instructional Fidelity errors, 330 Physical \& Integration Fidelity errors, 127 Background Fidelity errors, and 69 Object Omission errors (see Figure \ref{fig:piecharts}). Object Visual Fidelity errors are the most frequent type in most domains, Physical \& Integration Fidelity errors dominate in cosmetics, and Spatial \& Instructional Fidelity errors are common in furniture and cars. The per-domain percentages appear in the supplementary material (Section 3).


\begin{figure}[h]
\centering
\begin{minipage}{\linewidth}
\begin{lstlisting}[
    language=json,
    frame=single,
    numbers=none,
    basicstyle=\small\ttfamily,
    breaklines=true
]
"hallucination": 
{
  "objects": "Object mutation: The chair has been modified into a rocking chair.",
  "background": "",
  "position_logic": "Misplacement: The couch is not positioned against the wall.",
  "physical": "",
  "object_omission": ""
}
\end{lstlisting}
\end{minipage}
\caption{JSON with assigned hallucination class(es) and descriptions from exemplary sample from the VIGIL Benchmark Dataset.}
\label{fig:annotation_json}
\end{figure}

\subsection{Data Preparation}
Prompts for the recontextualization task were synthesized with Gemini 2.5 Flash. We provided the model with a system prompt constructed in accordance with the official Gemini guidelines, together with a set of reference images. The resulting prompts were inspected manually to verify their consistency with the input images and to confirm that no mismatches or semantic differences were introduced during automatic generation. The instructions with images were then passed to Gemini 2.5 Flash Image, selected as a representative multimodal system, to produce the final outputs and to reflect a realistic distribution of errors made by capable models. The hallucination classification and the explanatory descriptions were performed manually, without any involvement of language models during labeling.

\subsection{Description of Data Source}
All background and object images used in this work were obtained from publicly available online resources. For the clothing category, we used data from the Dress Code dataset \cite{morelli2022dresscode}. For the other categories, we collected high-resolution images from Kaggle and Unsplash (electronics), Douglas and Unsplash (cosmetics), and from targeted web crawling for furniture and cars.

\subsection{Taxonomy of Hallucination Types}
To assess hallucinations produced by generative recontextualization models, we define a taxonomy that decomposes visual and structural inconsistencies into five non-overlapping categories. The taxonomy was developed iteratively through a qualitative analysis of an initial subset of images, without a predefined scheme. Over several rounds of trial annotation, candidate categories were merged, split, and redefined until they applied consistently across the semantic domains. Each category isolates one failure mode related to object appearance, background preservation, spatial reasoning, physical integration, or object presence. The taxonomy separates localized artifacts from broader structural inconsistencies within the scene.

\subsubsection{Object Visual Fidelity}
This category captures hallucinations related exclusively to the visual appearance of the objects instructed to be inserted into the scene. The evaluation focuses on whether its identity matches the reference. Object Visual Fidelity captures deviations in texture, material, color, and geometric structure.

\subsubsection{Background Fidelity}
Background Fidelity isolates hallucinations affecting the surrounding environment instead of the inserted objects. This includes all contextual elements that are expected to remain unchanged, such as walls, floors, windows, lighting, and preexisting furniture. Violations occur when the background is altered, simplified, or replaced. 

\subsubsection{Spatial \& Instructional Fidelity}
This category captures failures related to spatial reasoning and compliance with high-level editing instructions. It concerns whether objects are placed in the correct locations, whether replacement or removal instructions are executed properly, and whether additional unintended objects are introduced. Appearance-related errors are excluded and handled by Object Visual Fidelity. 

\subsubsection{Physical \& Integration Fidelity}
This category evaluates whether the inserted objects are coherently integrated into the physical and visual structure of the scene. This category captures violations of real-world consistency, including lighting, shadows, perspective, and scale.

\subsubsection{Object Omission}
Object Omission refers to hallucinations in which the model fails to generate one or more objects that are explicitly required by the input instructions. Unlike other categories, this failure mode is defined by the complete absence of an expected object.

\subsection{Annotation Protocol \& Quality}
All samples were labeled with a multi-stage protocol. The dataset was first annotated by two independent human annotators. A third independent person then reviewed the samples to resolve discrepancies and to consolidate the final ground-truth annotations. Because our labels are descriptive, we measured inter-annotator agreement with an LLM-as-a-Judge framework (Gemini 2.5 Flash). The model decomposed the textual rationales from both primary annotators into distinct error instances and matched them. From these matches we computed the Positive Specific Agreement (PSA) and obtained a score of 0.8129, which indicates high consistency in identifying fine-grained visual inconsistencies.

\section{Methodology}
We describe the VIGIL pipeline for hallucination detection in recontextualization tasks. To our knowledge, this is the first method designed for fine-grained hallucination detection under multimodal consistency. The workflow decomposes verification into four axes, namely object-level fidelity, background consistency, physical plausibility, and spatial and instructional coherence. A diagram of the full pipeline and further details are provided in the supplementary material (Section 2).

\subsection{Object Extraction and Segmentation}
The process begins by identifying the entities that must be preserved. We use Qwen3.5-9B to parse the recontextualization prompt and extract the primary physical subjects. Once identified, we employ the Segment Anything Model 3 (SAM 3) to locate and mask these entities in both the reference and generated images. This provides the spatial coordinates and masks necessary for comparative analysis.

\subsection{Object Fidelity Verification}
For each segmented object, we generate high-dimensional embeddings using DINO v3-ViT-B/16. To determine correspondences between reference and generated objects, we compute a cosine similarity matrix and apply the Linear Sum Assignment method to solve the bipartite matching problem. The set of candidate pairs is restricted by semantic consistency. Only pairs sharing the same class label, as assigned by SAM3, are considered potential matches.

A pair is considered a valid match only if it satisfies the condition $S_{cos} > \tau$. If a reference object fails to find a valid match, it is flagged as an \textit{object omission} hallucination. Validated pairs are then passed back to the VLM (Qwen3.5-9B), which performs visual reasoning to detect \textit{object mutations} or \textit{reference bleeding} by comparing the segmented crops.

\subsection{Background Fidelity Verification}
To evaluate environmental integrity, we isolate the background by masking out all pixels that belong to the identified objects. To prevent the background evaluation from being biased by object-related artifacts such as contact shadows and reflections, we extend the mask by a margin. The margin is computed relative to the object's bounding box dimensions, so the exclusion zone scales with the object size and captures its peripheral influence. We use two methods.

\begin{itemize}
    \item \textbf{Direct comparative analysis} compares the masked reference and generated images to identify structural discrepancies or context swaps.
    \item \textbf{Difference-guided localization} computes the absolute pixel-wise difference between the masked backgrounds to identify regions of high variance. Bounding boxes are drawn around these regions, which guide the VLM to a targeted assessment of potential background hallucinations.
\end{itemize}

In both methods, the model is instructed to distinguish major structural errors from minor natural modifications, such as contact shadows or subtle lighting adjustments, that are required for realistic object integration.

\subsection{Physical Fidelity Verification}
Beyond identity and background integrity, the pipeline assesses the physical plausibility of the generated image at two granularities. At the \textbf{object level}, each matched reference and generated pair is presented to the VLM together with its surrounding visual context, isolated through an adaptive padding ratio relative to the object's bounding box. The model is prompted to detect physical anomalies such as inconsistent lighting and shadow direction, missing or geometrically implausible shadows, perspective and scale mismatches with the host scene, and visual artifacts introduced during insertion. At the \textbf{scene level}, the full generated image is examined for global physical incoherence, which captures violations that emerge only when objects are considered jointly with their environment, such as objects floating without contact with supporting surfaces, multiple light sources with mutually inconsistent directions, or unrealistic occlusion ordering. The two granularities operate at independent resolutions to balance computational cost against the level of detail required at each stage.

\subsection{Spatial and Instructional Verification}
The final axis addresses violations of the explicit instructional content of the recontextualization prompt, which the previous stages do not capture. We use the VLM to distill the prompt into a set of atomic constraints. Each constraint corresponds either to a \textit{positional relation} (e.g., \textit{``a chair placed to the right of the table''}) or to a \textit{replacement directive} (e.g., \textit{``replacing the smaller grey sofa''}). Constraints are derived only from explicit spatial or replacement wording. Lighting, shadow direction, and stylistic descriptors are excluded to prevent spurious flags. Each constraint is then evaluated individually against the generated image, which lets the VLM flag three failure types. Misplacement occurs when an object's position contradicts the constraint. Replacement failure occurs when the object that should have been substituted out remains visible alongside its replacement. Unwanted objects occur when additional foreground elements that are not specified in any constraint are introduced. Decomposing the prompt into independent constraints before verification, rather than asking the model to assess the scene holistically, improves the interpretability of the detected errors and the reliability of the VLM's judgement.

\section{Experiments \& Results}

To evaluate the proposed pipeline, we benchmarked its predictions against the ground-truth annotations in our dataset. The evaluation focuses on the ability of the system to identify and categorize hallucinations in the recontextualized outputs.

We evaluated three vision-language models (VLMs) acting as standalone hallucination detectors, namely Gemini 2.5 Flash, Qwen3.5-9B, and Gemma 3 12B IT. We selected Qwen3.5-9B and Gemma 3 12B IT because they are the largest models in their respective families that run on a single GPU in our setup. Qwen3.5-9B is also the model used inside our pipeline, which lets us measure its standalone performance.

As comparison baselines, we adapt two existing image-editing evaluators, VIEScore (SDIE protocol) and the agentic CIGEval. We run both on the same backbone as VIGIL (Qwen3.5-9B) for a fair comparison. Both methods emit only scalar quality scores with free-text rationales rather than per-category labels, so we classify these rationales with the same model into VIGIL's five hallucination categories and score them under the identical macro-F1 metric. Following VIGIL's protocol, all methods are evaluated leave-one-domain-out across the five product domains. For each held-out domain, the decision threshold $\tau$ (and, where applicable, the score margin) is calibrated on the remaining four domains by maximizing macro-F1, and only then applied to the held-out domain, so that no test-domain labels are ever used for tuning.

For each baseline model, we followed a fixed prompting protocol for a fair comparison.

\begin{itemize}
    \item \textbf{Prompt optimization} tailors the query instructions to the prompting guidelines of each model. We provide the Gemini 2.5 Flash baseline prompt in the supplementary material (Section 6).

    \item \textbf{Hallucination definitions} from the taxonomy used during our manual annotation, including descriptions of object distortions, missing entities, and background inconsistencies, are provided to each model.

    \item \textbf{Zero-shot detection} presents each model with the source images, the text prompt, and the generated output without task-specific training or fine-tuning, and asks it to identify all visible hallucinations.
\end{itemize}
The outputs of these models were compared against our annotations to determine how the multi-stage pipeline performs relative to the all-in-one detection approach of multimodal models.

\subsection{Evaluation Metrics}
We use two evaluation methods. The first is based on multi-label classification metrics for binary detection, and the second is LLM-as-a-Judge for semantic consistency.

\subsubsection{Multi-label Classification Performance}
We treat the detection task as a multi-label classification problem across the five categories of the taxonomy, namely Object Visual Fidelity, Background Fidelity, Spatial \& Instructional Fidelity, Physical \& Integration Fidelity, and Object Omission. The F1-score is computed for each hallucination type, and we report the macro-F1 across types, because the dataset is imbalanced both in error types (TN, TP, FP, FN) and in object categories (e.g., furniture, cars). Macro-averaging gives a more balanced evaluation across categories.

\subsubsection{LLM-as-a-Judge Semantic Evaluation}
Binary matching does not capture the semantic nuance of natural language descriptions, in particular when several distinct hallucinations are present in a single image. We therefore employ Gemini 2.5 Flash as a semantic judge that operates at the defect level. Instead of classifying the overall text alignment, the judge compares the set of hallucinations described by the pipeline against the human ground truth to quantify specific error instances. For each pair of descriptions, the model computes \textbf{True Positives (TP)}, the defects semantically present in both texts, \textbf{False Positives (FP)}, the defects produced by the pipeline but absent in the annotations, and \textbf{False Negatives (FN)}, the ground-truth defects missed by the pipeline. These defect-level counts are then used to compute F1-scores, so the metric reflects the ability of the system to isolate individual visual anomalies. To check the reliability of the judge and reduce the risk of meta-hallucinations, we conducted a manual blind audit on $100$ random instances. After prompt calibration, the judge reached a $92\%$ Exact Match Rate on the [TP, FP, FN] vector relative to human expert judgment.

\subsection{Results}
We calibrated the pipeline over four parameters. We varied the cosine similarity threshold used for object matching ($\tau \in \{0.1, 0.2, \ldots, 0.8\}$), the spatial margin applied to object masks during background verification ($\delta \in \{0.0, 0.1, 0.2\}$), the inclusion of difference-based ROI bounding boxes in the background prompt ($b \in \{\text{on}, \text{off}\}$), and the bounding-box padding ratio used to crop objects for the physical-fidelity stage ($\rho \in \{0.5, 1.0\}$).

The calibration was performed over our five categories, namely cars, clothing, cosmetics, furniture, and electronics. We followed a leave-one-category-out strategy. For each split, we ran a grid search on four categories and selected the best parameter set by the macro-F1 for the multi-label classification task. The selected configuration was then evaluated on the held-out category. We repeated this procedure for all five $4+1$ splits. The detailed results appear in the supplementary material (Section 1). Each cell in Table~\ref{tab: tab_binary} and Table~\ref{tab: tab_llm} reports a single macro-F1 for the held-out domain under this leave-one-domain-out protocol, so no per-cell variance is reported.

The results are summarized in Table~\ref{tab: tab_binary}. VIGIL achieves the best performance on furniture and outperforms all standalone VLM baselines, including Gemini 2.5 Flash. In the remaining categories the result is domain-dependent. VIGIL is competitive with Qwen3.5-9B and Gemma 3 12B IT but does not consistently outperform all baselines, and Gemini 2.5 Flash remains strongest on clothing and cars. This binary evaluation only measures whether a hallucination category is detected at all. To assess whether the detected errors are semantically correct and match the human annotations, we further use the LLM-as-a-Judge evaluation.

In the LLM-as-a-Judge setting, we considered only the parameter configurations previously selected as optimal for each category from the multi-label classification results. As shown in Table~\ref{tab: tab_llm}, VIGIL outperforms the open-source baselines in most categories and achieves the best non-Gemini result in clothing, furniture, cosmetics, and electronics, while Qwen3.5-9B performs better on cars. VIGIL also outperforms Gemma 3 12B IT across all categories, although Gemma is larger than the Qwen variant used inside our pipeline.

The two adapted baselines fall well below the VLM baselines and VIGIL in every domain in both Table~\ref{tab: tab_binary} and Table~\ref{tab: tab_llm}, and VIEScore records the lowest scores throughout. Their scalar scores, once reclassified into the five categories, carry little category-level signal, which supports the use of native per-category detection.

We underline the Gemini 2.5 Flash results because this model obtains the highest scores overall. These scores should be read with caution, because Gemini is also used as the LLM-as-a-Judge and may favor outputs produced in a similar style. The comparison with Qwen remains important, because this model is used several times within VIGIL. The results indicate that applying Qwen in a decomposed setting, where it solves focused subtasks, usually leads to stronger performance than using it as a standalone detector. The relatively low F1-scores in Table~\ref{tab: tab_llm} reflect the difficulty of detecting subtle hallucinations.

\begin{table}[h]
    \centering
    \caption{Macro-F1 for multi-label classification (higher is better), using the best parameter set on each test category. The remaining categories were used for calibration. Bold values indicate the best score among all non-Gemini methods. Underlined values indicate the second-best score among all non-Gemini methods.}
    \label{tab: tab_binary}
    \resizebox{\columnwidth}{!}{%
    \begin{tabular}{@{} lccccc @{}}
        \toprule
        \textbf{Method} & \textbf{Clothing} & \textbf{Furniture} & \textbf{Cosmetics} & \textbf{Electronics} & \textbf{Cars} \\
        \midrule
        Gemini 2.5 Flash & 0.4261 & 0.3696 & 0.2302 & 0.2890 & 0.4283 \\
        \midrule
        Qwen3.5-9B       & \underline{0.2927} & 0.4592 & \textbf{0.3455} & \underline{0.3046} & \underline{0.3747} \\
        Gemma 3 12B IT   & 0.2283 & \underline{0.4712} & 0.2932 & \textbf{0.3115} & 0.3743 \\
        CIGEval          & 0.1697 & 0.1845 & 0.1303 & 0.1845 & 0.2455 \\
        VIEScore         & 0.0348 & 0.0139 & 0.0034 & 0.0036 & 0.0952 \\
        VIGIL (ours)     & \textbf{0.3891} & \textbf{0.5064} & \underline{0.3442} & 0.3029 & \textbf{0.4154} \\
        \bottomrule
    \end{tabular}%
    }
\end{table}

\begin{table}[h]
    \centering
    \caption{Macro-F1 for the semantic LLM-as-a-Judge evaluation (higher is better), using the best parameter set selected for each test category from the multi-label classification calibration. Bold values indicate the best score among all non-Gemini methods. Underlined values indicate the second-best score among all non-Gemini methods. Because Gemini is also used as the judge, these results are reported only as a reference.}
    \label{tab: tab_llm}
    \resizebox{\columnwidth}{!}{%
    \begin{tabular}{@{} lccccc @{}}
        \toprule
        \textbf{Method} & \textbf{Clothing} & \textbf{Furniture} & \textbf{Cosmetics} & \textbf{Electronics} & \textbf{Cars} \\
        \midrule
        Gemini 2.5 Flash & 0.3185 & 0.2141 & 0.0911 & 0.1559 & 0.2724 \\
        \midrule
        Qwen3.5-9B       & \underline{0.1528} & \underline{0.1528} & \underline{0.0633} & \underline{0.0714} & \textbf{0.1683} \\
        Gemma 3 12B IT   & 0.0527 & 0.1175 & 0.0604 & 0.0427 & 0.0603 \\
        CIGEval          & 0.0756 & 0.0929 & 0.0415 & 0.0484 & 0.0977 \\
        VIEScore         & 0.0407 & 0.0103 & 0.0035 & 0.0022 & 0.0866 \\
        VIGIL (ours)     & \textbf{0.2455} & \textbf{0.1609} & \textbf{0.0902} & \textbf{0.0861} & \underline{0.1461} \\
        \bottomrule
    \end{tabular}%
    }
\end{table}

\subsection{Error Analysis}
To understand the remaining performance gaps, we performed a qualitative domain-level analysis of the categories in which the pipeline underperforms most, mainly Cars and Electronics. In Cars, most object-fidelity errors are localized text changes, and 67\% of annotated object mutations in this category involve license plate text distortion. Such errors preserve the overall vehicle identity, shape, and color but alter a small region that is visually non-salient for VLM comparison, so they are often missed. In Electronics, object identity is often determined by fine-grained details such as logos, screen contents, ports, keyboard layout, and small text. During recontextualization these details occupy only a small fraction of the generated image and may be blurred or downsampled in the extracted crop, which creates a scale mismatch between high-resolution reference objects and lower-resolution generated crops. Representative examples appear in the supplementary material (Section 4). These cases indicate that the main bottleneck in these domains is the lack of local verification rather than the overall decomposition of VIGIL. We also report an ablation of the SAM 3 segmentation stage, which measures how segmentation errors propagate to the downstream error categories, in the supplementary material (Section 5).

\section{Potential Applications}
The VIGIL framework is useful across several domains. In digital marketing, it verifies that product attributes remain constant during automated scene generation. In e-commerce, the pipeline supports virtual try-on systems by flagging garment mutations. The dataset is also a diagnostic tool for model developers, who can use it to identify whether weaknesses lie in spatial reasoning or in low-level feature preservation.
\section{Discussion and Limitations}
The VIGIL pipeline reaches strong category-specific F1-scores, but it has several limitations. The architecture combines SAM 3, DINO v3, and Qwen3.5-9B, which adds computational overhead. Model scaling is bounded by hardware availability. Larger variants could improve reasoning, but we use smaller models to fit the memory of a single A100 40 GB GPU. Dataset quality is also affected by technical factors. The initial image set was generated before the release of Gemini 3 Pro Image, so it has lower native resolution than current outputs. This produces a scale discrepancy during evaluation. When a generated object occupies only a small fraction of the frame, while in the reference image it occupies most of the pixels, the model compares low-resolution content against high-resolution content and lacks sufficient visual information for an accurate assessment. The evaluation framework also has limitations in semantic accuracy. When a generated object is not correctly paired with its reference counterpart, the background evaluation metrics become distorted. The model can also confuse similar objects, for example different products within a single cosmetics line whose visual features are nearly identical. Finally, the LLM-as-a-Judge approach is more informative than binary metrics, but it can still inherit biases from the underlying evaluator model.

Future work should focus on improving the pipeline's sensitivity to fine-grained inconsistencies — small logos, brand markings, and subtle textural details that are currently often missed because the cropped object resolution is insufficient for the VLM to discriminate. A complementary direction is a systematic study of alternative backbones inside VIGIL.

\section{Conclusion}
We introduced VIGIL, a framework for hallucination detection in multimodal image recontextualization. With a fine-grained taxonomy and a manually annotated benchmark dataset of 1,269 samples, we move from generic alignment scores to a structured diagnostic approach that isolates specific failure modes such as object mutations and background inconsistencies.

Our experiments show that decomposing the detection process into specialized modules for object fidelity and background consistency lets an open-source backbone match or exceed other open-source detectors and approach a much larger proprietary model, while it also returns an explanation for each error. The pipeline is strongest in domains that require high visual precision, which indicates the value of combining a discriminative model such as DINO v3 with a generative reasoner. General-purpose multimodal models are capable, but they perform better with structured, multi-stage guidance when identifying subtle visual hallucinations.

As generative models are adopted in e-commerce and digital advertising, automatic verification of visual integrity becomes important. The current approach targets semantic and visual fidelity, as well as geometric and physical reasoning. We intend the VIGIL benchmark and methodology to support further work on the reliability and explainability of large multimodal models.
\section{Statements and Declarations}
We gratefully acknowledge the Polish high-performance computing infrastructure PLGrid (HPC Center ACK Cyfronet AGH) for providing computer facilities and support within computational grant no. PLG/2025/018661.

{
    \small
    \bibliographystyle{ieeenat_fullname}
    \bibliography{main}
}

\clearpage
\appendix

\setcounter{page}{1}
\setcounter{section}{0}
\setcounter{figure}{0}
\setcounter{table}{0}

\renewcommand{\thesection}{\Alph{section}}
\renewcommand{\thefigure}{S\arabic{figure}}
\renewcommand{\thetable}{S\arabic{table}}

\twocolumn[{%
  \centering
  \LARGE \textbf{Supplementary Material \\ \vspace{0.5em} VIGIL: Tackling Hallucination Detection in Image Recontextualization} \par
  \vspace{2em}
}]

\section{Grid search results}\label{sec:appa}
Table \ref{tab:experiment_results} presents the results of a comprehensive grid search conducted to optimize the VIGIL pipeline's hyperparameters across different semantic categories. We calibrated the pipeline over four parameters. We varied the cosine similarity threshold used for object matching ($\tau \in \{0.1, 0.2, \ldots, 0.8\}$), the spatial margin applied to object masks during background verification ($\delta \in \{0.0, 0.1, 0.2\}$), the inclusion of difference-based ROI bounding boxes in the background prompt ($b \in \{\text{on}, \text{off}\}$), and the bounding-box padding ratio used to crop objects for the physical-fidelity stage ($\rho \in \{0.5, 1.0\}$).  

The configurations yielding the highest F1 scores for each category (highlighted in bold) were selected for final evaluation on the test split. Although the exact optimal hyperparameter sets vary slightly depending on the semantic domain, the overall performance of the VIGIL pipeline proves to be highly stable. As shown in Table 1, the F1 scores do not differ significantly and systematically oscillate around similar values across a wide range of configurations, provided that the cosine similarity threshold remains relatively low ($\tau \le 0.5$). Furthermore, variations in the spatial margin ($\delta$), ROI bounding boxes ($b$), and padding ratio ($\rho$) introduce only marginal fluctuations in the final metrics. This consistency demonstrates that our method is robust and maintains high accuracy across diverse tasks without the need for exhaustive, per-category manual tuning.

\begin{table*}[p]
\centering
\caption{Experiment Results: Binary Classification F1 Scores}
\fontsize{5.3pt}{6.3pt}\selectfont
\label{tab:experiment_results}
\begin{tabular}{rrrrrrrrr}
\toprule
\multicolumn{4}{c}{} & \multicolumn{5}{c}{Binary classification F1 score - all classes except} \\
\cmidrule(lr){5-9}
Threshold & Margin & Boxes & Padding & Cars & Clothes & Cosmetics & Electronics & Furniture \\
\midrule
0.1 & 0.0 & False & 0.5 & 0.4070 & 0.3985 & 0.4282 & 0.4338 & 0.3875 \\
0.1 & 0.0 & False & 1.0 & 0.4078 & 0.3996 & 0.4279 & 0.4346 & 0.3882 \\
0.1 & 0.0 & True & 0.5 & 0.4026 & 0.4004 & 0.4221 & 0.4287 & 0.3806 \\
0.1 & 0.0 & True & 1.0 & 0.4015 & 0.3995 & 0.4219 & 0.4277 & 0.3795 \\
0.1 & 0.1 & False & 0.5 & 0.4099 & 0.4010 & 0.4285 & 0.4359 & 0.3889 \\
0.1 & 0.1 & False & 1.0 & 0.4067 & 0.3999 & 0.4271 & 0.4331 & 0.3861 \\
0.1 & 0.1 & True & 0.5 & 0.4009 & 0.3981 & 0.4214 & 0.4274 & 0.3825 \\
0.1 & 0.1 & True & 1.0 & 0.4020 & 0.3990 & 0.4223 & 0.4288 & 0.3840 \\
0.1 & 0.2 & False & 0.5 & 0.4091 & 0.4019 & 0.4275 & 0.4351 & \textbf{0.3891} \\
0.1 & 0.2 & False & 1.0 & 0.4086 & 0.4016 & 0.4270 & 0.4344 & 0.3882 \\
0.1 & 0.2 & True & 0.5 & 0.4031 & 0.4033 & 0.4237 & 0.4300 & 0.3833 \\
0.1 & 0.2 & True & 1.0 & 0.4039 & 0.4037 & 0.4236 & 0.4307 & 0.3839 \\
0.2 & 0.0 & False & 0.5 & 0.4080 & 0.3982 & 0.4286 & 0.4366 & 0.3860 \\
0.2 & 0.0 & False & 1.0 & 0.4093 & 0.3994 & 0.4290 & 0.4380 & 0.3875 \\
0.2 & 0.0 & True & 0.5 & 0.4016 & 0.4082 & 0.4157 & 0.4268 & 0.3783 \\
0.2 & 0.0 & True & 1.0 & 0.4007 & 0.4074 & 0.4151 & 0.4257 & 0.3774 \\
0.2 & 0.1 & False & 0.5 & 0.4083 & 0.3982 & 0.4290 & 0.4373 & 0.3878 \\
0.2 & 0.1 & False & 1.0 & 0.4092 & 0.3984 & 0.4284 & 0.4373 & 0.3879 \\
0.2 & 0.1 & True & 0.5 & 0.4025 & 0.4100 & 0.4168 & 0.4283 & 0.3776 \\
0.2 & 0.1 & True & 1.0 & 0.4023 & 0.4097 & 0.4170 & 0.4282 & 0.3774 \\
0.2 & 0.2 & False & 0.5 & 0.4102 & 0.4009 & 0.4282 & 0.4379 & 0.3876 \\
0.2 & 0.2 & False & 1.0 & 0.4091 & 0.4012 & 0.4271 & 0.4369 & 0.3866 \\
0.2 & 0.2 & True & 0.5 & 0.4052 & \textbf{0.4138} & 0.4158 & 0.4304 & 0.3790 \\
0.2 & 0.2 & True & 1.0 & 0.4044 & 0.4122 & 0.4163 & 0.4298 & 0.3783 \\
0.3 & 0.0 & False & 0.5 & 0.4104 & 0.3977 & \textbf{0.4298} & 0.4383 & 0.3871 \\
0.3 & 0.0 & False & 1.0 & 0.4103 & 0.3972 & 0.4292 & 0.4377 & 0.3866 \\
0.3 & 0.0 & True & 0.5 & 0.4008 & 0.4031 & 0.4135 & 0.4246 & 0.3757 \\
0.3 & 0.0 & True & 1.0 & 0.3994 & 0.4023 & 0.4125 & 0.4230 & 0.3741 \\
0.3 & 0.1 & False & 0.5 & 0.4105 & 0.3966 & 0.4279 & 0.4371 & 0.3868 \\
0.3 & 0.1 & False & 1.0 & 0.4080 & 0.3975 & 0.4260 & 0.4353 & 0.3850 \\
0.3 & 0.1 & True & 0.5 & 0.4017 & 0.4060 & 0.4155 & 0.4263 & 0.3763 \\
0.3 & 0.1 & True & 1.0 & 0.4021 & 0.4069 & 0.4149 & 0.4268 & 0.3767 \\
0.3 & 0.2 & False & 0.5 & 0.4109 & 0.3993 & 0.4290 & 0.4382 & 0.3865 \\
0.3 & 0.2 & False & 1.0 & 0.4097 & 0.3988 & 0.4271 & 0.4366 & 0.3853 \\
0.3 & 0.2 & True & 0.5 & 0.4069 & 0.4103 & 0.4163 & 0.4308 & 0.3796 \\
0.3 & 0.2 & True & 1.0 & 0.4042 & 0.4088 & 0.4149 & 0.4275 & 0.3771 \\
0.4 & 0.0 & False & 0.5 & 0.4091 & 0.3959 & 0.4259 & 0.4369 & 0.3825 \\
0.4 & 0.0 & False & 1.0 & 0.4098 & 0.3965 & 0.4259 & 0.4376 & 0.3832 \\
0.4 & 0.0 & True & 0.5 & 0.4012 & 0.4037 & 0.4120 & 0.4264 & 0.3735 \\
0.4 & 0.0 & True & 1.0 & 0.4009 & 0.4038 & 0.4117 & 0.4261 & 0.3732 \\
0.4 & 0.1 & False & 0.5 & 0.4103 & 0.3966 & 0.4258 & 0.4378 & 0.3843 \\
0.4 & 0.1 & False & 1.0 & \textbf{0.4110} & 0.3979 & 0.4259 & \textbf{0.4391} & 0.3855 \\
0.4 & 0.1 & True & 0.5 & 0.4011 & 0.4063 & 0.4115 & 0.4266 & 0.3728 \\
0.4 & 0.1 & True & 1.0 & 0.4007 & 0.4044 & 0.4132 & 0.4261 & 0.3725 \\
0.4 & 0.2 & False & 0.5 & 0.4095 & 0.3979 & 0.4252 & 0.4370 & 0.3824 \\
0.4 & 0.2 & False & 1.0 & 0.4107 & 0.3989 & 0.4258 & 0.4385 & 0.3838 \\
0.4 & 0.2 & True & 0.5 & 0.4061 & 0.4082 & 0.4142 & 0.4312 & 0.3752 \\
0.4 & 0.2 & True & 1.0 & 0.4046 & 0.4083 & 0.4133 & 0.4301 & 0.3740 \\
0.5 & 0.0 & False & 0.5 & 0.4027 & 0.3926 & 0.4205 & 0.4318 & 0.3766 \\
0.5 & 0.0 & False & 1.0 & 0.4065 & 0.3928 & 0.4238 & 0.4353 & 0.3802 \\
0.5 & 0.0 & True & 0.5 & 0.3969 & 0.4001 & 0.4083 & 0.4233 & 0.3672 \\
0.5 & 0.0 & True & 1.0 & 0.3963 & 0.4000 & 0.4085 & 0.4225 & 0.3668 \\
0.5 & 0.1 & False & 0.5 & 0.4060 & 0.3949 & 0.4213 & 0.4344 & 0.3786 \\
0.5 & 0.1 & False & 1.0 & 0.4068 & 0.3943 & 0.4228 & 0.4352 & 0.3794 \\
0.5 & 0.1 & True & 0.5 & 0.3996 & 0.4042 & 0.4112 & 0.4243 & 0.3697 \\
0.5 & 0.1 & True & 1.0 & 0.4000 & 0.4048 & 0.4114 & 0.4245 & 0.3702 \\
0.5 & 0.2 & False & 0.5 & 0.4082 & 0.3968 & 0.4237 & 0.4369 & 0.3801 \\
0.5 & 0.2 & False & 1.0 & 0.4072 & 0.3958 & 0.4235 & 0.4356 & 0.3789 \\
0.5 & 0.2 & True & 0.5 & 0.4053 & 0.4081 & 0.4127 & 0.4292 & 0.3726 \\
0.5 & 0.2 & True & 1.0 & 0.4027 & 0.4076 & 0.4123 & 0.4266 & 0.3705 \\
0.6 & 0.0 & False & 0.5 & 0.3897 & 0.3807 & 0.4064 & 0.4182 & 0.3676 \\
0.6 & 0.0 & False & 1.0 & 0.3894 & 0.3812 & 0.4067 & 0.4180 & 0.3676 \\
0.6 & 0.0 & True & 0.5 & 0.3811 & 0.3767 & 0.3978 & 0.4083 & 0.3576 \\
0.6 & 0.0 & True & 1.0 & 0.3845 & 0.3804 & 0.4003 & 0.4121 & 0.3621 \\
0.6 & 0.1 & False & 0.5 & 0.3916 & 0.3812 & 0.4069 & 0.4192 & 0.3678 \\
0.6 & 0.1 & False & 1.0 & 0.3915 & 0.3809 & 0.4077 & 0.4188 & 0.3673 \\
0.6 & 0.1 & True & 0.5 & 0.3846 & 0.3797 & 0.4001 & 0.4123 & 0.3628 \\
0.6 & 0.1 & True & 1.0 & 0.3834 & 0.3791 & 0.3999 & 0.4110 & 0.3616 \\
0.6 & 0.2 & False & 0.5 & 0.3925 & 0.3826 & 0.4068 & 0.4193 & 0.3675 \\
0.6 & 0.2 & False & 1.0 & 0.3910 & 0.3824 & 0.4074 & 0.4192 & 0.3672 \\
0.6 & 0.2 & True & 0.5 & 0.3849 & 0.3820 & 0.4012 & 0.4127 & 0.3618 \\
0.6 & 0.2 & True & 1.0 & 0.3844 & 0.3815 & 0.4005 & 0.4123 & 0.3611 \\
0.7 & 0.0 & False & 0.5 & 0.3639 & 0.3656 & 0.3790 & 0.3889 & 0.3467 \\
0.7 & 0.0 & False & 1.0 & 0.3634 & 0.3646 & 0.3794 & 0.3880 & 0.3459 \\
0.7 & 0.0 & True & 0.5 & 0.3589 & 0.3643 & 0.3744 & 0.3833 & 0.3411 \\
0.7 & 0.0 & True & 1.0 & 0.3602 & 0.3657 & 0.3747 & 0.3846 & 0.3419 \\
0.7 & 0.1 & False & 0.5 & 0.3660 & 0.3668 & 0.3810 & 0.3908 & 0.3481 \\
0.7 & 0.1 & False & 1.0 & 0.3648 & 0.3663 & 0.3799 & 0.3899 & 0.3469 \\
0.7 & 0.1 & True & 0.5 & 0.3595 & 0.3653 & 0.3746 & 0.3847 & 0.3432 \\
0.7 & 0.1 & True & 1.0 & 0.3586 & 0.3643 & 0.3746 & 0.3838 & 0.3422 \\
0.7 & 0.2 & False & 0.5 & 0.3657 & 0.3680 & 0.3795 & 0.3906 & 0.3472 \\
0.7 & 0.2 & False & 1.0 & 0.3652 & 0.3675 & 0.3793 & 0.3898 & 0.3458 \\
0.7 & 0.2 & True & 0.5 & 0.3592 & 0.3649 & 0.3736 & 0.3830 & 0.3400 \\
0.7 & 0.2 & True & 1.0 & 0.3591 & 0.3666 & 0.3750 & 0.3845 & 0.3415 \\
0.8 & 0.0 & False & 0.5 & 0.3068 & 0.3285 & 0.3196 & 0.3297 & 0.3034 \\
0.8 & 0.0 & False & 1.0 & 0.3101 & 0.3318 & 0.3195 & 0.3323 & 0.3069 \\
0.8 & 0.0 & True & 0.5 & 0.3097 & 0.3307 & 0.3179 & 0.3310 & 0.3050 \\
0.8 & 0.0 & True & 1.0 & 0.3089 & 0.3311 & 0.3193 & 0.3309 & 0.3054 \\
0.8 & 0.1 & False & 0.5 & 0.3142 & 0.3335 & 0.3233 & 0.3368 & 0.3107 \\
0.8 & 0.1 & False & 1.0 & 0.3116 & 0.3331 & 0.3231 & 0.3344 & 0.3088 \\
0.8 & 0.1 & True & 0.5 & 0.3092 & 0.3322 & 0.3184 & 0.3316 & 0.3057 \\
0.8 & 0.1 & True & 1.0 & 0.3080 & 0.3319 & 0.3193 & 0.3311 & 0.3061 \\
0.8 & 0.2 & False & 0.5 & 0.3112 & 0.3330 & 0.3207 & 0.3338 & 0.3074 \\
0.8 & 0.2 & False & 1.0 & 0.3111 & 0.3332 & 0.3205 & 0.3333 & 0.3071 \\
0.8 & 0.2 & True & 0.5 & 0.3085 & 0.3319 & 0.3170 & 0.3301 & 0.3051 \\
0.8 & 0.2 & True & 1.0 & 0.3083 & 0.3317 & 0.3188 & 0.3309 & 0.3059 \\
\bottomrule
\normalsize
\end{tabular}
\end{table*}

\section{Multi-stage detection pipeline}
\begin{figure*}[t]
  \centering
  \includegraphics[width=0.78\textwidth]{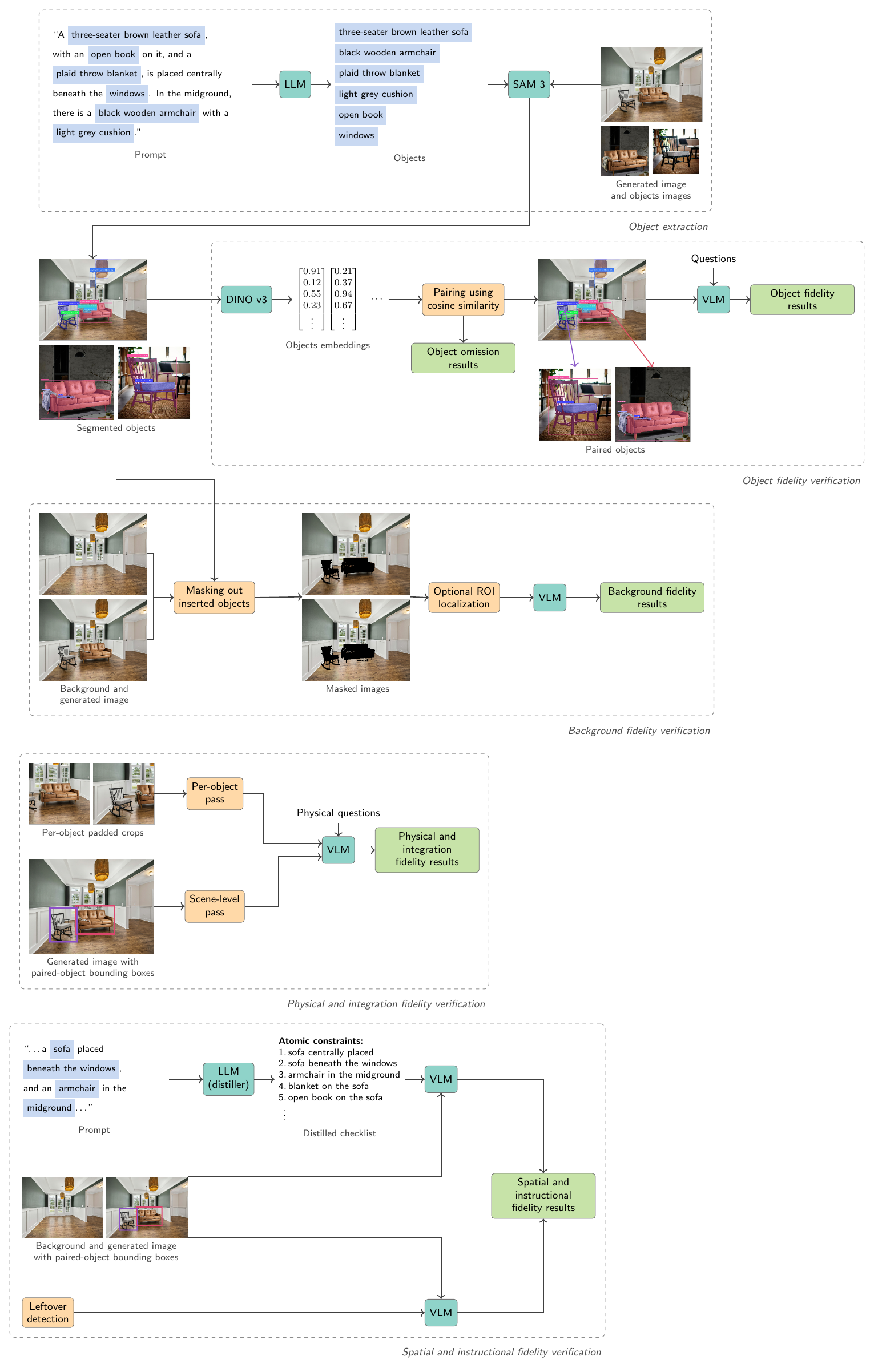}
  \caption{Architecture of the multi-stage detection pipeline. Entity extraction, segmentation
and embedding produce a paired representation of the sample, which is then consumed by
ﬁve category-speciﬁc veriﬁcation modules.}
\label{fig: pipeline_schema}
  \label{fig:pipeline}
\end{figure*}

To supplement the methodological description provided in the main paper, Figure \ref{fig: pipeline_schema} presents a comprehensive visual trace of the VIGIL pipeline applied to an exemplary generation task. Rather than reiterating the underlying architecture, this diagram illustrates the practical data flow of a single instance passing through our multi-stage framework. It explicitly visualizes how intermediate representations are constructed at each step—such as isolated object masks from SAM 3, masked backgrounds for environmental consistency evaluation, and adaptive crops for physical reasoning. By tracing the inputs from the initial prompt parsing down to the final VLM-generated diagnostic outputs, the figure demonstrates how the complex task of multimodal hallucination detection is successfully decomposed into specialized, interpretable subtasks.

\section{Dataset statistics}
Table \ref{tab:domain-cat} details the distribution of hallucination types across the five semantic domains, revealing that the frequency of specific errors is highly domain-dependent. Object Visual Fidelity is the most prevalent issue overall, affecting a significant majority of Furniture (73.3\%) and Cars (52.2\%) samples. However, certain domains exhibit distinct failure modes; for instance, Physical \& Integration errors dominate the Cosmetics category (51.1\%), while Spatial \& Instructional inconsistencies are particularly common in both Cars (52.2\%) and Furniture (51.4\%). Conversely, Background alterations and Object Omissions remain consistently the least frequent hallucination types across all evaluated datasets, with omissions dropping to near zero in categories like Cosmetics and Electronics.

\begin{table*}[htbp]
    \centering
    \caption{Percentage of samples in each domain that exhibit at least one error of each top-level category. Rows correspond to the five semantic domains of the dataset, and columns correspond to the five categories of the taxonomy.}
    \label{tab:domain-cat}
    \begin{tabular*}{\textwidth}{@{\extracolsep{\fill}} lccccc @{}}
        \toprule
        \textbf{Category} & \textbf{Objects} & \textbf{Background} & \textbf{Spatial\&Instructional} & \textbf{Physical\&Integration} & \textbf{Object Omission} \\
        \midrule
        Furniture       & 73.3 & 10.5 & 51.4 & 41.9 & 8.6 \\
        Clothing        & 45.7 & 12.5 & 29.8 &  3.0 & 10.5 \\
        Cosmetics       & 23.8 & 15.2 & 28.3 & 51.1 &  0.0 \\
        Electronics     & 46.6 &  1.9 & 16.3 & 42.0 &  1.1 \\
        Cars            & 52.2 &  8.3 & 52.2 & 25.6 &  2.8 \\
        \bottomrule
    \end{tabular*}%
\end{table*}

\section{Domain-specific error analysis}
\begin{figure*}[t]
    \centering
    \includegraphics[width=\linewidth]{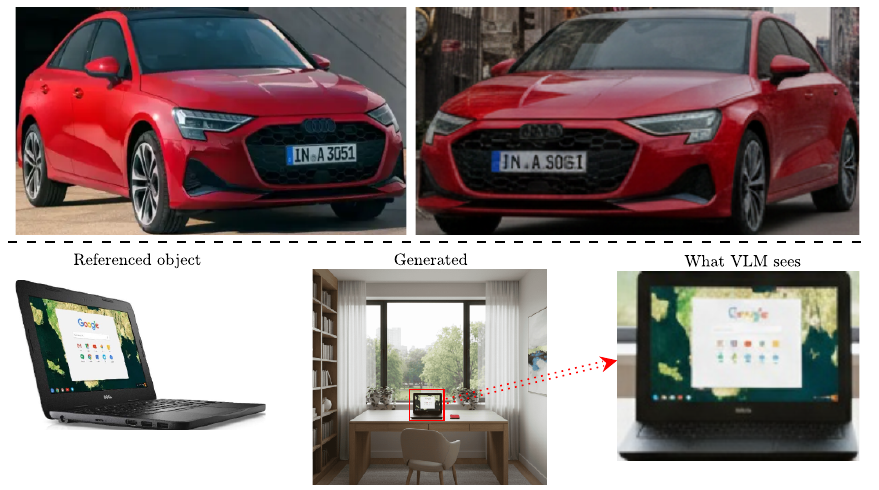}
    \caption{
    Representative domain-specific failure cases. In the top row, for the Cars category, the generated vehicle preserves the global appearance of the reference car, but the license plate text is distorted, which illustrates a localized object-fidelity error. In the bottom row, for the Electronics category, the generated laptop loses fine-grained identity cues such as screen content, logo or text, and keyboard details, because of the low effective resolution of the generated crop.
    }
    \label{fig:domain_error_analysis}
\end{figure*}

Figure~\ref{fig:domain_error_analysis} shows representative failure cases for the Cars and Electronics domains.

\section{SAM ablation}
SAM 3 operates at the first stage of VIGIL and governs which objects are extracted for downstream verification. A segmentation error at this stage propagates forward. Missed objects produce false negatives in the object omission and object fidelity checks, while spurious detections introduce false positives in the background consistency module.

To quantify this propagation, we manually annotated 20 randomly sampled images per product category (100 in total). For each image, we enumerated all entities mentioned in the generation prompt, treated them as class labels, and drew bounding boxes around each entity in both the generated and reference images. These annotations served as ground truth for evaluating SAM's detection accuracy.

SAM's predictions were then compared against these annotations using the standard object detection metrics mAP@0.5 and mAP@0.95. In parallel, each image was assigned to one of two groups. An image is \textit{SAM-correct} if every predicted bounding box was matched to a ground-truth box with IoU $\geq 0.5$ and a correct class label, and \textit{SAM-incorrect} otherwise. This yielded 52 correct and 45 incorrect samples. Table~\ref{tab:sam_map} reports per-category detection rates and mAP scores for both groups.

To test whether SAM segmentation quality has a statistically significant effect on downstream pipeline errors, we counted the pipeline's per-image true positive (TP), false positive (FP), false negative (FN), and F1 scores across the background, object omission, and objects error categories for each group. We then applied the Mann--Whitney U test \cite{mannwhitney1947}, a non-parametric rank-based test whose null hypothesis states that the two independent samples are drawn from the same distribution, that SAM segmentation quality does not affect downstream pipeline error metrics. The $p$-values were corrected for multiple comparisons with the Holm--Bonferroni procedure. The results are reported in Table~\ref{tab:sam_mw}.

The tests reveal a consistent propagation pattern. SAM segmentation quality has a highly significant effect on false positive rates in both the background ($p_{\text{adj}} < 0.001$) and object omission ($p_{\text{adj}} < 0.001$) categories, and the SAM-incorrect group shows substantially higher median FP counts (0.800 and 0.911, respectively) than the SAM-correct group (0.288 and 0.058). When SAM fails to isolate objects accurately, the pipeline misattributes object-level errors as spurious background inconsistencies or phantom omissions. The downstream impact on background F1 is also significant ($p_{\text{adj}} = 0.041$), and the SAM-correct group reaches a markedly higher median score (0.283 vs.\ 0.011). True positive and false negative rates are not significantly affected in any category, which indicates that SAM errors inflate noise rather than cause genuine hallucinations to be missed.

\begin{table*}[htbp]
    \centering
    \caption{SAM 3 detection performance per error category for the manually annotated subset. Det.\% is the fraction of images in each group where the pipeline correctly detected at least one instance of the hallucination type. C is SAM-correct and I is SAM-incorrect.}
    \label{tab:sam_map}
    \resizebox{\textwidth}{!}{%
    \begin{tabular}{@{} lcccccc @{}}
        \toprule
        \textbf{Category} & \textbf{Det.\% (C)} & \textbf{Det.\% (I)} & \textbf{mAP@0.5 (C)} & \textbf{mAP@0.5 (I)} & \textbf{mAP@0.95 (C)} & \textbf{mAP@0.95 (I)} \\
        \midrule
        Background       & 71\% & 31\% & 0.701 & 0.712 & 0.725 & 0.718 \\
        Object Omission  & 94\% & 49\% & 0.742 & 0.942 & 0.690 & 0.797 \\
        Objects          & 42\% & 31\% & 0.546 & 0.423 & 0.611 & 0.500 \\
        Physical\&Integration         &  4\% &  4\% & 0.464 & 0.038 & 0.500 & 0.071 \\
        Spatial\&Instructional & 33\% & 24\% & 0.526 & 0.327 & 0.607 & 0.425 \\
        \bottomrule
    \end{tabular}%
    }
\end{table*}

\begin{table}[htbp]
    \centering
    \caption{Mann--Whitney U test comparing per-image pipeline error metrics between the SAM-correct ($n_C$) and SAM-incorrect ($n_I$) groups. The $p$-values are Holm--Bonferroni adjusted. For F1, only images with at least one positive instance are included ($n$ varies). Significance levels are $^{***}p<0.001$, $^{*}p<0.05$, and \textit{ns} not significant.}
    \label{tab:sam_mw}
    \resizebox{\columnwidth}{!}{%
    \begin{tabular}{@{} lcccc @{}}
        \toprule
        \textbf{Metric} & \textbf{Med.\ (C)} & \textbf{Med.\ (I)} & \textbf{$p_{\text{adj}}$} & \textbf{Sig.} \\
        \midrule
        Background TP       & 0.115 & 0.022 & 0.336 & \textit{ns} \\
        Background FP       & 0.288 & 0.800 & 0.000 & $^{***}$ \\
        Background FN       & 0.058 & 0.089 & 0.884 & \textit{ns} \\
        Background F1       & 0.283 & 0.011 & 0.041 & $^{*}$ \\
        \midrule
        Object Omission TP  & 0.000 & 0.022 & 0.511 & \textit{ns} \\
        Object Omission FP  & 0.058 & 0.911 & 0.000 & $^{***}$ \\
        Object Omission FN  & 0.000 & 0.022 & 0.511 & \textit{ns} \\
        Object Omission F1  & 0.000 & 0.042 & 0.884 & \textit{ns} \\
        \midrule
        Objects TP          & 0.288 & 0.333 & 0.977 & \textit{ns} \\
        Objects FP          & 0.731 & 1.133 & 0.511 & \textit{ns} \\
        \bottomrule
    \end{tabular}%
    }
\end{table}

\section{Exemplary prompt used for Gemini 2.5 Flash baseline}\label{sec:prompts}

\begin{figure*}[t]
  \begin{promptbox}
    \textbf{You are an image re-contextualization hallucination inspector.} \\
    \textbf{You will be given:} \\
    - An instruction prompt (what the generator was asked to do). \\
    - A background image (background\_image). \\
    - Reference object image(s) (one or two) - object1\_image, object2\_image. \\
    - The generated image (generated\_image) - result of re-contextualization.
    
    \textbf{Your task:} compare the generated\_image to the references and to the instruction, and produce \textbf{SHORT SEMANTIC DESCRIPTIONS} (1-3 sentences each) answering the following categories. If no issues or hallucinations are detected in a category, return an empty string ("") for that category.
    
    \textbf{Categories:}
    
    \textbf{1) objects:} Object Visual Fidelity - texture/shape/color identity mismatches, mutations, identity loss, reference bleeding. Example: "Feature Mutation: sofa's color changed from dark green to black; Identity Loss: inserted cabinet is metallic vs wicker."
    
    \textbf{2) background:} Background Fidelity - background mutations, background detail loss, context swap. Example: "Background Mutation: wall color changed; Context Swap: bedroom replaced by living room."
    
    \textbf{3) position\_logic:} Spatial and Instructional Fidelity - misplacement of inserted object, replacement failure (the item that should have been replaces remain in the image). Example: "Misplacement: The green chair is on the right side of the bed instead of the left side."
    
    \textbf{4) physical:} Physical and Integration Fidelity - lighting/shadow incoherence, perspective/scale issues, artifacts (unnatural phenomena). Example: "Shadow Incoherence: The is no shadow under the inserted chair."
    
    \textbf{5) object\_omission:} Object Omission - missing required objects from the instruction that should have been pasted from object image. Example: "Object Omission: green cabinet missing."
    
    \textbf{Instruction Prompt:} \{instruction\_prompt\}
    
    Analyze the provided images now.
  \end{promptbox}
  \caption{Exemplary system prompt used for the Gemini 2.5 Flash baseline. The prompt assigns a specific role and breaks the task into five distinct categories with semantic definitions, ensuring consistent evaluation.}
  \label{fig:gemini_prompt}
\end{figure*}

Prompt \ref{fig:gemini_prompt} follows multimodal design principles by assigning a specific role and breaking the task into five distinct categories. This structure directs the model’s attention to specific visual elements and provides clear semantic definitions for errors, ensuring consistent evaluation without the need for visual examples.

\section{Exemplary prompt used for the LLM-as-a-Judge semantic evaluation}

\begin{figure*}[t]
  \begin{promptbox}
    \textbf{You are an expert auditor evaluating AI hallucination detection.} \\
    
    \textbf{Hallucination Type:} \{hallucination\_type\} \\
    
    \textbf{Human Ground Truth (list of real errors):} \\
    "\{ground\_truth\_text\}" \\
    
    \textbf{AI Pipeline Output (list of detected errors):} \\
    "\{pipeline\_text\}" \\
    
    \textbf{Task:} \\
    Break down BOTH descriptions into individual distinct errors/issues and count: \\
    
    \textbf{1. TP (True Positives):} How many specific errors from Ground Truth did the AI correctly detect? \\
    - Synonyms count as matches (e.g., "missing leg" == "leg not visible") \\
    - Similar descriptions of the same error count as TP \\
    
    \textbf{2. FN (False Negatives):} How many specific errors from Ground Truth did the AI MISS completely? \\
    - Count each distinct error that's in Ground Truth but NOT in AI output \\
    
    \textbf{3. FP (False Positives):} How many NEW errors did the AI report that are NOT in Ground Truth? \\
    - Count each distinct error that's in AI output but NOT in Ground Truth \\
    
    \textbf{Rules:} \\
    - Treat each distinct object/issue as a separate error instance \\
    -- If multiple errors are described in one sentence, count them separately \\
    - Be precise and conservative in counting \\
    
    \textbf{Examples:} \\
    - Ground Truth = "Object omission: missing red car and blue truck" has 2 errors, not 1 \\
    - Ground Truth = "Object mutation: the back and sides of the car do not correspond to the reference" has 1 error \\
    - For instance: \\
    \hspace*{0.5cm} - "Object mutation: The jeans have different details" \\
    \hspace*{0.5cm} - "Object mutation: The jeans show incorrect stitching patterns and colors" \\
    Both descriptions refer to the same visual inconsistency and should be treated as a single TP. \\
    
    \textbf{Output ONLY valid JSON in this exact format:} \\
    \{ \\
    \hspace*{0.5cm} "tp": <integer>, \\
    \hspace*{0.5cm} "fn": <integer>, \\
    \hspace*{0.5cm} "fp": <integer> \\
    \}
  \end{promptbox}
  \caption{System prompt used for the LLM-as-a-Judge semantic evaluation. The prompt instructs the model to extract and count True Positives (TP), False Positives (FP), and False Negatives (FN) based on a direct comparison between the human ground-truth descriptions and the AI pipeline predictions.}
  \label{fig:auditor_prompt}
\end{figure*}

Prompt \ref{fig:auditor_prompt} presents the system prompt used for the LLM-as-a-Judge semantic evaluation. The instructions direct the model to compare human ground-truth annotations with the AI pipeline's predictions by isolating individual hallucination instances. By enforcing strict counting rules for True Positives (TP), False Positives (FP), and False Negatives (FN), this prompt ensures a robust, semantic assessment of complex textual descriptions that standard string matching would miss.

\section{Results per hallucination category}
As detailed in Table \ref{tab:results_per_cat}, the fine-grained semantic evaluation across our five hallucination categories reveals distinct performance patterns and highlights the strengths of a decomposed pipeline.

VIGIL demonstrates a clear advantage in detecting localized visual inconsistencies. Notably, our method achieves the highest F1-scores in the Objects (0.2627) and Background (0.1449) categories, outperforming both the open-source VLM baselines (Qwen3.5-9B, Gemma 3) and the adapted agentic evaluators (CIGEval, VIEScore). Importantly, VIGIL even surpasses the much larger proprietary model (Gemini 2.5 Flash) in these two areas. This validates our architectural design: by leveraging explicit object extraction (SAM 3) and discriminative feature matching (DINO v3), VIGIL is able to capture subtle textural, structural, and identity mutations that global, standalone VLM assessments frequently miss.

In contrast, Gemini 2.5 Flash dominates in the Spatial \& Instructional (0.4866) and Object Omission (0.5495) categories. This reflects the strong holistic scene understanding and advanced spatial reasoning inherent to massive proprietary models. However, it should be noted that these specific scores may be partially inflated due to Gemini's dual role as the semantic judge in this evaluation protocol, potentially favoring outputs formatted in its native reasoning style.

Finally, the results show that existing holistic evaluators like VIEScore and CIGEval struggle significantly with producing accurate, fine-grained semantic descriptions, yielding F1-scores near zero in several categories. Furthermore, the Physical \& Integration category remains a significant challenge across the board for all methods. This suggests that detecting subtle lighting, shadow, and perspective inconsistencies is an open problem that requires deeper physical reasoning beyond what current 2D visual analysis and VLM prompting can reliably provide.

\begin{table*}[htbp]
    \centering
    \caption{Macro-F1 for the semantic LLM-as-a-Judge evaluation (higher is better) per hallucination category. Bold values indicate the best score among all non-Gemini methods. Because Gemini is also used as the judge, these results are reported only as a reference.}
    \label{tab:results_per_cat}
    \begin{tabular*}{\textwidth}{@{\extracolsep{\fill}} lccccc @{}}
        \toprule
        \textbf{Method} & \textbf{Objects} & \textbf{Background} & \textbf{Spatial\&Instructional} & \textbf{Physical\&Integration} & \textbf{Object Omission} \\
        \midrule
        Gemini 2.5 Flash & 0.1422 & 0.0442 & 0.4866 & 0.1347 & 0.5495 \\
        \midrule
        Qwen3.5-9B       & 0.1781 & 0.0284 & 0.1979 & 0.0532 & \textbf{0.2158} \\
        Gemma 3 12B IT   & 0.0578 & 0.0110 & 0.1132 & \textbf{0.1535} & 0.0233 \\
        CIGEval          & 0.1369 & 0.0066 & 0.0972 & 0.0145 & 0.1022 \\
        VIEScore         & 0.0074 & 0.0000 & 0.0000 & 0.0250 & 0.1750 \\
        VIGIL (ours)     & \textbf{0.2627} & \textbf{0.1449} & \textbf{0.2379} & 0.0603 & 0.1028 \\
        \bottomrule
    \end{tabular*}%
\end{table*}

\end{document}